\documentclass{article}

\usepackage{microtype}
\usepackage{graphicx}
\usepackage{subfigure}
\usepackage{booktabs} 
\usepackage{latexsym}
\usepackage{amsfonts}
\usepackage{mathrsfs}
\usepackage{amssymb,amscd}
\usepackage[all]{xy}
\usepackage{amsmath}
\usepackage{fancyhdr}
\usepackage{fancybox}
\usepackage{xcolor}
\usepackage{array}
\usepackage{makecell}
\usepackage{multirow}
\usepackage{url}

\usepackage{hyperref}

\usepackage{amsthm}
\newtheorem{theorem}{Theorem}[]
\newtheorem{lemma}[]{Lemma}[]

\newtheorem{definition}[]{Definition}[]

\newtheorem*{problem}{Problem Statement}

\newcommand{\mO}[1]{\mathcal{O}\left(#1\right)}

\usepackage[accepted]{innf2021}

\icmltitlerunning{Universal Approximation of Residual Flows in Maximum Mean Discrepancy}

\begin{document}

\twocolumn[
\icmltitle{Universal Approximation of Residual Flows in Maximum Mean Discrepancy}


\begin{icmlauthorlist}
\icmlauthor{Zhifeng Kong}{ucsd}
\icmlauthor{Kamalika Chaudhuri}{ucsd}

\end{icmlauthorlist}

\icmlaffiliation{ucsd}{Department of Computer Science and Engineering, University of California San Diego, CA, USA}

\icmlcorrespondingauthor{Zhifeng Kong}{z4kong@eng.ucsd.edu}
\icmlcorrespondingauthor{Kamalika Chaudhuri}{kamalika@cs.ucsd.edu}

\icmlkeywords{Residual flows, normalizing flows, Maximum Mean Discrepancy, universal approximation, expressiveness}

\vskip 0.3in
]

\printAffiliationsAndNotice{}
\begin{abstract}
    Normalizing flows are a class of flexible deep generative models that offer easy likelihood computation. Despite their empirical success, there is little theoretical understanding of their expressiveness. In this work, we study residual flows, a class of normalizing flows composed of Lipschitz residual blocks. We prove residual flows are universal approximators in maximum mean discrepancy. We provide upper bounds on the number of residual blocks to achieve approximation under different assumptions. 
\end{abstract}

\section{Introduction}\label{sec: introduction}

Normalizing flows are a class of generative models that learn an invertible function to transform a predefined source distribution into a complex target distribution \citep{tabak2010density, tabak2013family, rezende2015variational}.
One category of normalizing flows called residual flows use residual networks \citep{he2016deep} to construct the transformation \citep{rezende2015variational,van2018sylvester,behrmann2019invertible,chen2019residual}. These models have shown great success in complicated real-world tasks. 

However, to ensure invertibility, these models apply additional Lipschitz constraints to each residual block. Under these strong constraints, how expressive these models are remains an open question. Formally, can they approximate certain target distributions to within any small error?

In this paper, we carry out a theoretical analysis on the expressive power of residual flows. We prove there exists a residual flow $F$ that achieves universal approximation in the mean maximum discrepancy (MMD, \cite{gretton2012kernel}) metric. Formally, given a target distribution, we provide upper bounds on the number of residual blocks in $F$ such that applying $F$ over the source distribution can approximate the target distribution in squared MMD (see \eqref{eq: goal}).


Although residual networks are universal approximators \citep{lin2018resnet}, the proof of approximation uses a non-invertible construction and therefore does not apply to residual flows. This reflects the main difficulty in analyzing residual flows: under strong Lipschitz and invertibility constraints, they become a very restricted function class. As an illustration, take the set of piecewise constant functions. Classical real analysis shows that piecewise constant functions can approximate any Lebesgue-integrable function and therefore any probability density. However, the invertible subset of all piecewise constant functions is the empty set! Consequently, this universal approximation result does not apply to normalizing flows. This difficulty leads to many negative results for normalizing flows: they are either unable to express or find it hard to approximate certain functions \citep{zhang2019approximation,koehler2020representational,pmlr-v108-kong20a}. 

To tackle this problem, we adopt a new construction that satisfies the strong Lipschitz constraints in \citet{behrmann2019invertible}. Specifically, we construct the residual blocks by multiplying a small $\epsilon$ to a pre-specified Lipschitz function. Therefore, as long as $\epsilon$ is small enough, the strong Lipschitz constraints are satisfied. We then analyze the following quantity: how much can the MMD be reduced if a new residual block is appended? 
%
Since this quantity is a function of $\epsilon$, we can analyze its Taylor expansion. With a first-order analysis and under mild conditions, we show there is an $F$ with $\Theta\left(\frac1\delta\left(\log\frac1\delta\right)^2\right)$ residual blocks that achieves \eqref{eq: goal} (see \textbf{Theorem} \ref{thm: first order}), where $\delta$ is the ratio between the final squared MMD and the initial squared MMD. With a second-order analysis and under more conditions, we show there is a shallower $F$ with only $\Theta\left(\log\frac1\delta\right)$ residual blocks that achieves \eqref{eq: goal} (see \textbf{Theorem} \ref{thm: second order}). 

To sum up, we show residual flows are universal approximators in MMD under certain assumptions and provide explicit bounds on the number of residual blocks.

\section{Related Work}\label{sec: related work}
The classic universal approximation theory for fully connected or residual neural networks in the function space are widely studied \citep{cybenko1989approximation, hornik1989multilayer, hornik1991approximation, montufar2014number, telgarsky2015representation, lu2017expressive, hanin2017universal, raghu2017expressive, lin2018resnet}. However, these results do not generalize to residual flows \citep{rezende2015variational,van2018sylvester,behrmann2019invertible,chen2019residual} for two reasons. First, the approximation theory for normalizing flows analyzes how well they can transform between \textit{distributions}, rather than their ability to approximate a target function in the \textit{function space}. Despite that $L^p$ universality in the function space may lead to distributional universality for triangular flows \citep{teshima2020coupling}, there is no similar results for non-triangular flows including residual flows. Second, the classic results do not consider the invertibility or the Lipschitz constraints of the neural networks, which greatly restrict the expressive power.

There are also universal approximation results for Lipschitz networks \citep{anil2019sorting, cohen2019universal, tanielian2020approximating}. These results are related because in this work, we assume the expressive power of each Lipschitz residual block is large. However, these results only apply to functions defined on compact sets. Because compact sets are bounded, it is ``easier'' to satisfy the Lipschitz constraints. It is not trivial to extend their results to functions defined on $\mathbb{R}^d$. 

Concerning the expressive power of generative networks, there are prior works showing feed-forward generator networks can approximate certain distributions \citep{lee2017ability,bailey2018size,lu2020universal,perekrestenko2020constructive}. However, the results are again based on non-invertible constructions, so they do not apply to normalizing flows.

In the literature of normalizing flows, there are universal approximation results for several models including autoregressive flows \citep{germain2015made,kingma2016improved,papamakarios2017masked,huang2018neural,jaini2019sum}, coupling flows \citep{teshima2020coupling,koehler2020representational}, and augmented normalizing flows \citep{zhang2019approximation,huang2020solving} \footnote{In an augmented normalizing flow, there is an auxiliary random variable concatenated with the data, so the transformations operate on a higher dimensional space.}. There is also a continuous-time generalization of normalizing flows called neural ODEs \citep{chen2018neural,dupont2019augmented} with a universal approximation result \citep{zhang2019approximation}. We do not consider these flows in this paper. In addition, \cite{muller2020space} suggests residual networks can approximate neural ODEs, but the invertibility is again not considered in this case. 

On the expressive power of residual flows, all existing theoretical analysis present negative results for these models \citep{zhang2019approximation,koehler2020representational,pmlr-v108-kong20a}. These results indicate residual flows are either unable to express certain functions, or unable to approximate certain distributions even with large depths. Compared to these results, our paper presents positive results for standard residual flows: given a source distribution $q$, they can approximate a target distribution $p$ in the MMD metric \citep{gretton2012kernel} under certain conditions. We provide explicit upper bounds on the number of residual blocks (see \textbf{Theorem} \ref{thm: first order} and \textbf{Theorem} \ref{thm: second order}). 

\section{Preliminaries}\label{sec: preliminaries}
We first define the maximum mean discrepancy (MMD) metric between distributions below.

\begin{definition}[MMD, \cite{gretton2012kernel}]\label{def: MMD}
Let $q,p$ be two distributions on $\mathbb{R}^d$. Then,
\begin{equation}\label{eq: MMD kernel version}
\begin{array}{rl}
    \mathrm{MMD}(q,p)^2 = 
    & \mathbb{E}_{z,z'\sim q} K(z,z') + \mathbb{E}_{x,x'\sim p} K(x,x') \\
    & - 2\cdot\mathbb{E}_{z\sim q,x\sim p} K(z,x)
\end{array}
\end{equation}
for some kernel function $K(\cdot,\cdot)$. Let $\phi:\mathbb{R}^d\rightarrow\mathbb{R}^{d_{\phi}}$ be the feature map associated with $K$: $K(x,z)=\phi(x)^{\top}\phi(z)$, where we assume $d_{\phi}<\infty$. Then, the squared MMD can be simplified as
\begin{equation}\label{eq: MMD feature version}
    \mathrm{MMD}(q,p)^2=\|\mathbb{E}_{z\sim q}\phi(z)-\mathbb{E}_{x\sim p}\phi(x)\|_2^2.
\end{equation}
\end{definition}

Next, we define a residual flow as a composition of invertible layers parameterized as $\mathbf{Id}+f$, where $\mathbf{Id}$ is the identity map and $f$ is $\frac12$-Lipschitz\footnote{According to the fixed-point theorem, $\mathbf{Id}+f$ is invertible as long as the Lipschitz constant of $f$ is strictly less than 1. For algebraic convenience, we restrict the Lipschitz constant to be at most $\frac12$.}. The class of residual flows include planar flows \citep{rezende2015variational}, Sylvester flows \citep{van2018sylvester}, and the more general invertible residual networks \citep{behrmann2019invertible,chen2019residual}. In these models every $f_i$ is parameterized as a certain kind of fully-connected neural network. Since the expressive power of ($\frac12$-)Lipschitz neural networks on $\mathbb{R}^d$ remains an open problem, in this paper we assume every $f_i$ can be selected as any $\frac12$-Lipschitz function. Formally, we make the following definition.

\begin{definition}[Residual flows]\label{def: flow}
The set of $N$-block residual flows is defined as
\begin{equation}\label{eq: flow}
\begin{array}{rl}
    \mathcal{F}_N = 
    & \left\{(\mathbf{Id}+f_N)\circ\cdots\circ(\mathbf{Id}+f_1):\right. \\
    & \left. ~~~~~~~~~~~~~~~~~~~~~~~~ \text{each } f_i\text{ is }\frac12\text{-Lipschitz}\right\}.
\end{array}
\end{equation}
\end{definition}

Now we state the main problem. Let $q_{\mathrm{source}}$ and $p_{\mathrm{target}}$ be two distributions on $\mathbb{R}^d$, where $q_{\mathrm{source}}$ is the source distribution and $p_{\mathrm{target}}$ is the target distribution. We aim to answer the following problem in this paper.

\begin{problem}
Let $\delta>0$ be a small number. For any pair of distributions $q_{\mathrm{source}}$ and $p_{\mathrm{target}}$ on $\mathbb{R}^d$ satisfying $\mathrm{MMD}(q_{\mathrm{source}},p_{\mathrm{target}})<\infty$, does there exist an $N$ and $F\in\mathcal{F}_N$ such that 
\begin{equation}\label{eq: goal}
    \mathrm{MMD}(F\#q_{\mathrm{source}},p_{\mathrm{target}})^2 \leq \delta\cdot\mathrm{MMD}(q_{\mathrm{source}},p_{\mathrm{target}})^2,
\end{equation}
where $F\#q$ refers to the distribution obtained by applying $F$ over $q$?
\end{problem}

In this paper, we prove existence of such $F$ with a loose bound on $N$ using first-order analysis under mild assumptions (see Section \ref{sec: first order results}), and provide a tighter bound on $N$ using second-order analysis under more assumptions (see Section \ref{sec: second order results}).

\section{A Bound with First-Order Analysis}\label{sec: first order results}
In this section, we show under mild conditions, there exists a residual flow $F$ with $N=\Theta\left(\frac1\delta\left(\log\frac1\delta\right)^2\right)$ residual blocks that achieves \eqref{eq: goal}. The idea is to show that a single residual block can reduce the squared MMD by a certain fraction, so $F$ is obtained by stacking an enough number of these residual blocks. To begin with, we make the follow definition.

\begin{definition}\label{def: delta}
For distributions $q$, $p$, and a $\frac12$-Lipschitz function $f$, we define the improvement of the squared MMD by $\mathbf{Id}+f$ as
\begin{equation}\label{eq: delta}
    \Delta(q,p;f)=\mathrm{MMD}(q,p)^2 - \mathrm{MMD}((\mathbf{Id}+f)\#q,p)^2.
\end{equation}
\end{definition}

Then, if $\Delta(q,p;f) > 0$, the residual block $\mathbf{Id}+f$ is helpful for reducing the squared MMD. It is straightforward to see that $\sup\{\Delta(q,p;f)\text{: }f\text{ is }\frac12\text{-Lipschitz}\}\geq0$. In order to construct an $f$ that has a large $\Delta(q,p;f)$, we choose $f=\hat{f}_{\epsilon}$ defined below. 

\begin{definition}\label{def: f}
    Define $\psi(p,q)=(\mathbb{E}_{x\sim p}-\mathbb{E}_{x\sim q})\phi(x)$, $g(z)=\psi(p,q)^{\top}\phi(z)$, and $\hat{f}_{\epsilon}=\epsilon\cdot\nabla g$, where $\epsilon>0$. Then, $\mathrm{MMD}(q,p)=\|\psi(p,q)\|$. In addition, $\hat{f}_{\epsilon}(z)=\epsilon J_{\phi}(z)\psi(p,q)$, where $J_{\phi}$ is the Jacobian matrix of $\phi$.
\end{definition}

Then, according to \eqref{eq: MMD feature version} and \eqref{eq: delta},
\begin{equation}\begin{array}{rl}
    \Delta(q,p;\hat{f}_{\epsilon}) =
    & ~~~ \mathbb{E}_{z\sim q,x\sim q} \phi(z)^{\top}\phi(x)  \\
    & - \mathbb{E}_{z\sim q,x\sim q}\phi(z+\hat{f}_{\epsilon}(z))^{\top}\phi(x+\hat{f}_{\epsilon}(x)) \\
    & + 2\cdot \mathbb{E}_{z\sim q,x\sim p} \phi(z+\hat{f}_{\epsilon}(z))^{\top}\phi(x) \\
    & - 2\cdot \mathbb{E}_{z\sim q,x\sim p} \phi(z)^{\top}\phi(x) .
\end{array}\end{equation}
Note that $\Delta(q,p;\hat{f}_{\epsilon})$ is a function of $\epsilon$. We then analyze the first-order Taylor expansion of $\Delta(q,p;\hat{f}_{\epsilon})$ at $\epsilon=0^+$, denoted as $\Delta_1(q,p;\hat{f}_{\epsilon})$. Then, $\Delta(q,p;\hat{f}_{\epsilon}) = \Delta_1(q,p;\hat{f}_{\epsilon}) + \mO{\epsilon^2}$. With some arithmetic, we have
\begin{equation}\label{eq: delta_1 explicit}
    \Delta_1(q,p;\hat{f}_{\epsilon}) = 2\psi(p,q)^{\top} \mathbb{E}_{z\sim q}\phi(z+\hat{f}_{\epsilon}(z)).
\end{equation}
We have the following bound on $\Delta_1(q,p;\hat{f}_{\epsilon})$.

\begin{lemma}\label{lemma: bound Delta 1}
If $d_{\phi}<\infty$, and the minimum singular value $\sigma_{\min}(J_{\phi}(z))\geq\sqrt{ b}>0$ holds for any $z\in\mathbb{R}^d$, then 
\begin{equation}
    \Delta_1(q,p;\hat{f}_{\epsilon}) \geq 2\epsilon b\cdot\mathrm{MMD}(q,p)^2.
\end{equation}
\end{lemma}

Since $\Delta(q,p;\hat{f}_{\epsilon})=\Delta_1(q,p;\hat{f}_{\epsilon})+\mO{\epsilon^2}$, when $\epsilon$ is small, the residual block $\mathbf{Id}+\hat{f}_{\epsilon}$ can indeed reduce the squared MMD by a certain fraction ($\geq 2\epsilon b$). Next, as we require $f=\hat{f}_{\epsilon}$ to be $\frac12$-Lipschitz, we show under certain conditions the Lipschitz constant of $\hat{f}_{\epsilon}$ is $\mO{\epsilon}$ in the following lemma.

\begin{lemma}\label{lemma: lipschitz constant of f}
If for any $z\in\mathbb{R}^d$, the Lipschitz constant of each element in $J_{\phi}(z)$ is no more than a universal constant $L_{\mathrm{Jac}}$, then 
\begin{equation}
    \mathrm{Lip}(\hat{f}_{\epsilon})\leq\sqrt{d\cdot d_{\phi}}L_{\mathrm{Jac}}\mathrm{MMD}(q,p)\cdot \epsilon.
\end{equation}
\end{lemma}

With these tools, we can construct an $F\in\mathcal{F}_N$ that achieves \eqref{eq: goal} in the following theorem.

\begin{theorem}\label{thm: first order}
Under the conditions of \textbf{Lemma} \ref{lemma: bound Delta 1} and \textbf{Lemma} \ref{lemma: lipschitz constant of f}, there exists an $F\in\mathcal{F}_N$ with $N=\Theta\left(\frac1\delta\left(\log\frac1\delta\right)^2\right)$ that achieves \eqref{eq: goal}. 
\end{theorem}

The proof is deferred to Section \ref{appendix: thm 1}. The main idea in the proof is to construct each $f_i$ iteratively based on $f_1$ through $f_{i-1}$, so that adding this residual block can reduce the squared MMD by a certain fraction as indicated in \textbf{Lemma} \ref{lemma: bound Delta 1}. The bound is obtained by carefully balancing $\epsilon$, $\delta$, and $N$.

\section{A Tighter Bound with Second-Order Analysis}\label{sec: second order results}

In this section, we show under a few additional assumptions, there exists a much smaller $N=\mO{\log\frac1\delta}$ and $F\in\mathcal{F}_N$ such that $F$ achieves \eqref{eq: goal}. The idea is to bound the second-order remainder of the Taylor expansion of $\Delta(q,p;\hat{f}_{\epsilon})$: $\Delta_2(q,p;\hat{f}_{\epsilon})=\Delta(q,p;\hat{f}_{\epsilon})-\Delta_1(q,p;\hat{f}_{\epsilon})=\mO{\epsilon^2}$. Once $\Delta_2(q,p;\hat{f}_{\epsilon})$ is explicitly bounded we can pick a small constant $\epsilon$ for every residual block \footnote{In \textbf{Theorem} \ref{thm: first order}, the $\epsilon$ for each residual block is related to $\delta$ in order to eliminate the effect by the unknown second-order terms. Here $\epsilon$ is independent with $\delta$.} so $\Delta(q,p;\hat{f}_{\epsilon})$ is lower bounded by a universal constant times $\mathrm{MMD}(q,p)^2$. This then yields the $\mO{\log\frac1\delta}$ bound for $N$. Now, we provide an explicit bound on $\Delta_2(q,p;\hat{f}_{\epsilon})$ in the following lemma.

\begin{lemma}\label{lemma: bound Delta 2}
Let $B,C,L_{\mathrm{feat}}$ be positive constants. If for any $z\in\mathbb{R}^d$, the maximum singular value $\sigma_{\max}(J_{\phi}(z))\leq\sqrt{B}$, the absolute value of any eigenvalue $|\lambda(\nabla^2\phi_i(z))|\leq C$ for any $1\leq i\leq d_{\phi}$, and $\phi$ is $L_{\mathrm{feat}}$-Lipschitz, then 
\begin{equation}
\begin{array}{rl}
    |\Delta_2(q,p;\hat{f}_{\epsilon})| 
    & \leq \epsilon^2\cdot \mathrm{MMD}(q,p)^2 \cdot B \cdot \left( B + \text{ }^{\textcolor{white}{0}}\right.\\
    & ~~~~\left.\|\psi(p,q)\|\sqrt{d_{\phi}}C(1+\epsilon L_{\mathrm{feat}}\sqrt{B})\right).
\end{array}
\end{equation}
\end{lemma}

Given this explicit bound on $\Delta_2(q,p;\hat{f}_{\epsilon})$, we can pick a small $\epsilon$ such that $|\Delta_2(q,p;\hat{f}_{\epsilon})|\leq\frac12 \Delta_1(q,p;\hat{f}_{\epsilon})$ so that $\Delta(q,p;\hat{f}_{\epsilon})\geq\frac12 \Delta_1(q,p;\hat{f}_{\epsilon})$. Once this lower bound on $\Delta(q,p;\hat{f}_{\epsilon})$ is achieved, the squared MMD is multiplied by at most a universal constant less than $1$ when the new residual block $\mathbf{Id}+\hat{f}_{\epsilon}$ is added. We formalize the result in the following theorem.

\begin{theorem}\label{thm: second order}
Under the conditions of \textbf{Lemma} \ref{lemma: bound Delta 1}, \textbf{Lemma} \ref{lemma: lipschitz constant of f}, and \textbf{Lemma} \ref{lemma: bound Delta 2}, there exists an $F\in\mathcal{F}_N$ with $N=\Theta\left(\log\frac1\delta\right)$ that achieves \eqref{eq: goal}.
\end{theorem}

The proof is deferred to Section \ref{appendix: thm 2}. The main idea in the proof is to construct each $f_i$ in a similar way as in \textbf{Theorem} \ref{thm: first order}, but $\epsilon$ is selected as a universal constant according to \textbf{Lemma} \ref{lemma: bound Delta 2}.

\section{Conclusions}\label{sec: conclusion}
Normalizing flows are a class of flexible deep generative models that offers easy likelihood computation. Despite their empirical success, there is little theoretical understanding on whether they are universal approximators in transforming between probability distributions. In this work, we prove residual flows are indeed universal approximators in maximum mean discrepancy. Upper bounds on the number of residual blocks to achieve approximation are provided. Under mild conditions, we show  $\Theta\left(\frac1\delta\left(\log\frac1\delta\right)^2\right)$ residual blocks can achieve \eqref{eq: goal} (see \textbf{Theorem} \ref{thm: first order}). Under more conditions, we show as few as $\Theta\left(\log\frac1\delta\right)$ residual blocks can achieve \eqref{eq: goal} (see \textbf{Theorem} \ref{thm: second order}). 

There are a large number of open problems. One extension is to build universal approximation theory for residual flows in more general probability metrics such as the integral probability metrics \cite{muller1997integral} and the $f$-divergences \cite{csiszar2004information}. Another direction is to extend the proposed universal approximation theory to other classes of normalizing flows such as autoregressive flows. A final open problem is to look at normalizing flows in real-world applications, and analyze their expressive power under practical assumptions.  

\section*{Acknowledgement}
We thank Songbai Yan for helpful feedback.

\bibliographystyle{apalike}
\bibliography{bib.bib}

\begin{thebibliography}{}

\bibitem[Anil et~al., 2019]{anil2019sorting}
Anil, C., Lucas, J., and Grosse, R. (2019).
\newblock Sorting out lipschitz function approximation.
\newblock In {\em International Conference on Machine Learning}, pages
  291--301.

\bibitem[Bailey and Telgarsky, 2018]{bailey2018size}
Bailey, B. and Telgarsky, M.~J. (2018).
\newblock Size-noise tradeoffs in generative networks.
\newblock In {\em Advances in Neural Information Processing Systems}, pages
  6489--6499.

\bibitem[Behrmann et~al., 2019]{behrmann2019invertible}
Behrmann, J., Grathwohl, W., Chen, R.~T., Duvenaud, D., and Jacobsen, J.-H.
  (2019).
\newblock Invertible residual networks.
\newblock In {\em International Conference on Machine Learning}, pages
  573--582.

\bibitem[Chen et~al., 2019]{chen2019residual}
Chen, R.~T., Behrmann, J., Duvenaud, D.~K., and Jacobsen, J.-H. (2019).
\newblock Residual flows for invertible generative modeling.
\newblock In {\em Advances in Neural Information Processing Systems}, pages
  9916--9926.

\bibitem[Chen et~al., 2018]{chen2018neural}
Chen, T.~Q., Rubanova, Y., Bettencourt, J., and Duvenaud, D.~K. (2018).
\newblock Neural ordinary differential equations.
\newblock In {\em Advances in neural information processing systems}, pages
  6571--6583.

\bibitem[Cohen et~al., 2019]{cohen2019universal}
Cohen, J.~E., Huster, T., and Cohen, R. (2019).
\newblock Universal lipschitz approximation in bounded depth neural networks.
\newblock {\em arXiv preprint arXiv:1904.04861}.

\bibitem[Csisz{\'a}r and Shields, 2004]{csiszar2004information}
Csisz{\'a}r, I. and Shields, P.~C. (2004).
\newblock Information theory and statistics: A tutorial.

\bibitem[Cybenko, 1989]{cybenko1989approximation}
Cybenko, G. (1989).
\newblock Approximation by superpositions of a sigmoidal function.
\newblock {\em Mathematics of control, signals and systems}, 2(4):303--314.

\bibitem[Dupont et~al., 2019]{dupont2019augmented}
Dupont, E., Doucet, A., and Teh, Y.~W. (2019).
\newblock Augmented neural odes.
\newblock {\em arXiv preprint arXiv:1904.01681}.

\bibitem[Germain et~al., 2015]{germain2015made}
Germain, M., Gregor, K., Murray, I., and Larochelle, H. (2015).
\newblock Made: Masked autoencoder for distribution estimation.
\newblock In {\em ICML}.

\bibitem[Gretton et~al., 2012]{gretton2012kernel}
Gretton, A., Borgwardt, K.~M., Rasch, M.~J., Sch{\"o}lkopf, B., and Smola, A.
  (2012).
\newblock A kernel two-sample test.
\newblock {\em The Journal of Machine Learning Research}, 13(1):723--773.

\bibitem[Hanin, 2017]{hanin2017universal}
Hanin, B. (2017).
\newblock Universal function approximation by deep neural nets with bounded
  width and relu activations.
\newblock {\em arXiv preprint arXiv:1708.02691}.

\bibitem[He et~al., 2016]{he2016deep}
He, K., Zhang, X., Ren, S., and Sun, J. (2016).
\newblock Deep residual learning for image recognition.
\newblock In {\em Proceedings of the IEEE conference on computer vision and
  pattern recognition}, pages 770--778.

\bibitem[Hornik, 1991]{hornik1991approximation}
Hornik, K. (1991).
\newblock Approximation capabilities of multilayer feedforward networks.
\newblock {\em Neural networks}, 4(2):251--257.

\bibitem[Hornik et~al., 1989]{hornik1989multilayer}
Hornik, K., Stinchcombe, M., and White, H. (1989).
\newblock Multilayer feedforward networks are universal approximators.
\newblock {\em Neural networks}, 2(5):359--366.

\bibitem[Huang et~al., 2020]{huang2020solving}
Huang, C.-W., Dinh, L., and Courville, A. (2020).
\newblock Solving ode with universal flows: Approximation theory for flow-based
  models.
\newblock In {\em ICLR 2020 Workshop on Integration of Deep Neural Models and
  Differential Equations}.

\bibitem[Huang et~al., 2018]{huang2018neural}
Huang, C.-W., Krueger, D., Lacoste, A., and Courville, A.~C. (2018).
\newblock Neural autoregressive flows.
\newblock In {\em ICML}.

\bibitem[Jaini et~al., 2019]{jaini2019sum}
Jaini, P., Selby, K.~A., and Yu, Y. (2019).
\newblock Sum-of-squares polynomial flow.
\newblock In {\em International Conference on Machine Learning}, pages
  3009--3018.

\bibitem[Kingma et~al., 2016]{kingma2016improved}
Kingma, D.~P., Salimans, T., Jozefowicz, R., Chen, X., Sutskever, I., and
  Welling, M. (2016).
\newblock Improved variational inference with inverse autoregressive flow.
\newblock In {\em Advances in neural information processing systems}, pages
  4743--4751.

\bibitem[Koehler et~al., 2020]{koehler2020representational}
Koehler, F., Mehta, V., and Risteski, A. (2020).
\newblock Representational aspects of depth and conditioning in normalizing
  flows.
\newblock {\em arXiv preprint arXiv:2010.01155}.

\bibitem[Kong and Chaudhuri, 2020]{pmlr-v108-kong20a}
Kong, Z. and Chaudhuri, K. (2020).
\newblock The expressive power of a class of normalizing flow models.
\newblock In Chiappa, S. and Calandra, R., editors, {\em Proceedings of the
  Twenty Third International Conference on Artificial Intelligence and
  Statistics}, volume 108 of {\em Proceedings of Machine Learning Research},
  pages 3599--3609, Online. PMLR.

\bibitem[Lee et~al., 2017]{lee2017ability}
Lee, H., Ge, R., Ma, T., Risteski, A., and Arora, S. (2017).
\newblock On the ability of neural nets to express distributions.
\newblock In {\em Conference on Learning Theory}, pages 1271--1296.

\bibitem[Lin and Jegelka, 2018]{lin2018resnet}
Lin, H. and Jegelka, S. (2018).
\newblock Resnet with one-neuron hidden layers is a universal approximator.
\newblock In {\em Advances in Neural Information Processing Systems}, pages
  6169--6178.

\bibitem[Lu and Lu, 2020]{lu2020universal}
Lu, Y. and Lu, J. (2020).
\newblock A universal approximation theorem of deep neural networks for
  expressing distributions.
\newblock {\em arXiv preprint arXiv:2004.08867}.

\bibitem[Lu et~al., 2017]{lu2017expressive}
Lu, Z., Pu, H., Wang, F., Hu, Z., and Wang, L. (2017).
\newblock The expressive power of neural networks: A view from the width.
\newblock In {\em Advances in neural information processing systems}, pages
  6231--6239.

\bibitem[Montufar et~al., 2014]{montufar2014number}
Montufar, G.~F., Pascanu, R., Cho, K., and Bengio, Y. (2014).
\newblock On the number of linear regions of deep neural networks.
\newblock In {\em Advances in neural information processing systems}, pages
  2924--2932.

\bibitem[M{\"u}ller, 1997]{muller1997integral}
M{\"u}ller, A. (1997).
\newblock Integral probability metrics and their generating classes of
  functions.
\newblock {\em Advances in Applied Probability}, pages 429--443.

\bibitem[M{\"u}ller, 2020]{muller2020space}
M{\"u}ller, J. (2020).
\newblock On the space-time expressivity of resnets.
\newblock In {\em ICLR 2020 Workshop on Integration of Deep Neural Models and
  Differential Equations}.

\bibitem[Papamakarios et~al., 2017]{papamakarios2017masked}
Papamakarios, G., Pavlakou, T., and Murray, I. (2017).
\newblock Masked autoregressive flow for density estimation.
\newblock In {\em Advances in Neural Information Processing Systems}, pages
  2338--2347.

\bibitem[Perekrestenko et~al., 2020]{perekrestenko2020constructive}
Perekrestenko, D., M{\"u}ller, S., and B{\"o}lcskei, H. (2020).
\newblock Constructive universal high-dimensional distribution generation
  through deep relu networks.
\newblock {\em arXiv preprint arXiv:2006.16664}.

\bibitem[Raghu et~al., 2017]{raghu2017expressive}
Raghu, M., Poole, B., Kleinberg, J., Ganguli, S., and Dickstein, J.~S. (2017).
\newblock On the expressive power of deep neural networks.
\newblock In {\em Proceedings of the 34th International Conference on Machine
  Learning-Volume 70}, pages 2847--2854. JMLR. org.

\bibitem[Rezende and Mohamed, 2015]{rezende2015variational}
Rezende, D. and Mohamed, S. (2015).
\newblock Variational inference with normalizing flows.
\newblock In {\em International Conference on Machine Learning}, pages
  1530--1538.

\bibitem[Tabak and Turner, 2013]{tabak2013family}
Tabak, E.~G. and Turner, C.~V. (2013).
\newblock A family of nonparametric density estimation algorithms.
\newblock {\em Communications on Pure and Applied Mathematics}, 66(2):145--164.

\bibitem[Tabak et~al., 2010]{tabak2010density}
Tabak, E.~G., Vanden-Eijnden, E., et~al. (2010).
\newblock Density estimation by dual ascent of the log-likelihood.
\newblock {\em Communications in Mathematical Sciences}, 8(1):217--233.

\bibitem[Tanielian et~al., 2020]{tanielian2020approximating}
Tanielian, U., Sangnier, M., and Biau, G. (2020).
\newblock Approximating lipschitz continuous functions with groupsort neural
  networks.
\newblock {\em arXiv preprint arXiv:2006.05254}.

\bibitem[Telgarsky, 2015]{telgarsky2015representation}
Telgarsky, M. (2015).
\newblock Representation benefits of deep feedforward networks.
\newblock {\em arXiv preprint arXiv:1509.08101}.

\bibitem[Teshima et~al., 2020]{teshima2020coupling}
Teshima, T., Ishikawa, I., Tojo, K., Oono, K., Ikeda, M., and Sugiyama, M.
  (2020).
\newblock Coupling-based invertible neural networks are universal
  diffeomorphism approximators.
\newblock {\em Advances in Neural Information Processing Systems}, 33.

\bibitem[Van Den~Berg et~al., 2018]{van2018sylvester}
Van Den~Berg, R., Hasenclever, L., Tomczak, J.~M., and Welling, M. (2018).
\newblock Sylvester normalizing flows for variational inference.
\newblock In {\em 34th Conference on Uncertainty in Artificial Intelligence
  2018, UAI 2018}, pages 393--402. Association For Uncertainty in Artificial
  Intelligence (AUAI).

\bibitem[Zhang et~al., 2019]{zhang2019approximation}
Zhang, H., Gao, X., Unterman, J., and Arodz, T. (2019).
\newblock Approximation capabilities of neural ordinary differential equations.
\newblock {\em arXiv preprint arXiv:1907.12998}, 2(4).

\end{thebibliography}

\newpage
\onecolumn
\appendix

\section{Omitted Proofs}\label{sec: proof}

\subsection{Proof of \textbf{Lemma} \ref{lemma: bound Delta 1}}
\begin{proof}
According to \eqref{eq: delta_1 explicit} and the chain rule,
\[\begin{array}{rl}
    \Delta_1(q,p;\hat{f}_{\epsilon})
    & \displaystyle = 2\cdot \mathbb{E}_{z\sim q} \left(\psi(p,q)^{\top}J_{\phi}(z)^{\top}\hat{f}_{\epsilon}(z)\right) \\
    & \displaystyle = 2\epsilon\cdot \mathbb{E}_{z\sim q} \left(\psi(p,q)^{\top}J_{\phi}(z)^{\top}\nabla g(z)\right) \\
    & \displaystyle = 2\epsilon\cdot \mathbb{E}_{z\sim q} \left(\psi(p,q)^{\top}J_{\phi}(z)^{\top}J_{\phi}(z)\psi(p,q)\right) \\
    & \displaystyle \geq 2\epsilon\cdot \min_{z\in\mathbb{R}^d} \left(\psi(p,q)^{\top} J_{\phi}(z)^{\top}J_{\phi}(z)\psi(p,q)\right) \\
    \displaystyle \left(\mathrm{MMD}(q,p)^2=\|\psi(p,q)\|^2\right)
    & \displaystyle \geq 2\epsilon\cdot \mathrm{MMD}(q,p)^2 \min_{z\in\mathbb{R}^d} \lambda_{\min}\left(J_{\phi}(z)^{\top}J_{\phi}(z)\right) \\
    & \displaystyle \geq 2\epsilon\cdot \mathrm{MMD}(q,p)^2 \min_{z\in\mathbb{R}^d} \sigma_{\min}^2(J_{\phi}(z)) \\
    & \displaystyle \geq 2\epsilon b\cdot \mathrm{MMD}(q,p)^2.
\end{array}\]
\end{proof}

\subsection{Proof of \textbf{Lemma} \ref{lemma: lipschitz constant of f}}
\begin{proof}
For any $x,y\in\mathbb{R}^d$,
\[\begin{array}{rl}
    \displaystyle \frac{\|\hat{f}_{\epsilon}(y)-\hat{f}_{\epsilon}(x)\|}{\|y-x\|} 
    & \displaystyle = \frac{\epsilon\sqrt{\sum_{i=1}^d\left(\sum_{j=1}^{d_{\phi}} (J_{\phi}(y)-J_{\phi}(x))_{ij}\psi(p,q)_j\right)^2}}{\|y-x\|} \\
    & \displaystyle \leq \frac{\epsilon\sqrt{\sum_{i=1}^d\left(\sum_{j=1}^{d_{\phi}} L_{\mathrm{Jac}}\|y-x\|\psi(p,q)_j\right)^2}}{\|y-x\|} \\
    & \displaystyle \leq \epsilon\sqrt{d}L_{\mathrm{Jac}}\|\psi(p,q)\|_1 \\
    & \displaystyle \leq \epsilon\sqrt{d\cdot d_{\phi}}L_{\mathrm{Jac}}\|\psi(p,q)\|_2 \\
    & \displaystyle = \epsilon\sqrt{d\cdot d_{\phi}}L_{\mathrm{Jac}}\mathrm{MMD}(q,p). \\
\end{array}\]
Therefore, by taking the supreme over the left-hand-side, we have the Lipschitz constant of $\hat{f}_{\epsilon}$ is upper bounded by the right-hand-side.
\end{proof}

\subsection{Proof of \textbf{Theorem} \ref{thm: first order}}\label{appendix: thm 1}
\begin{proof}
Let $r>0$ and $\epsilon = r/N$. Define
\[D_n(r)=\mathrm{MMD}((\mathbf{Id}+f_n)\circ\cdots\circ(\mathbf{Id}+f_1)\#q_{\mathrm{source}},p_{\mathrm{target}})^2\]
where each 
\[f_i(z)=\epsilon J_{\phi}(z)\psi(p_{\mathrm{target}},(\mathbf{Id}+f_{i-1})\circ\cdots\circ(\mathbf{Id}+f_1)\#q_{\mathrm{source}}).\]
Note that each $f_i$ is exactly the $\hat{f}_{\epsilon}$ in \textbf{Definition} \ref{def: f} for
$q=(\mathbf{Id}+f_{i-1})\circ\cdots\circ(\mathbf{Id}+f_1)\#q_{\mathrm{source}}$ and $p=p_{\mathrm{target}}$.

By \textbf{Lemma} \ref{lemma: bound Delta 1}, 
\[D_n(r) \leq \left(1-2b\frac rN + \mO{\frac{r^2}{N^2}}\right) D_{n-1}(r).\]
Therefore,
\[\begin{array}{rl}
    D_N(r) 
    & \displaystyle \leq \prod_{n=1}^N\left(1-2b\frac rN + \mO{\frac{r^2}{N^2}}\right)\mathrm{MMD}(q_{\mathrm{source}},p_{\mathrm{target}})^2 \\
    & \displaystyle \leq \left(e^{-2br}+\mO{\frac{r^2}{N}}\right) \mathrm{MMD}(q_{\mathrm{source}},p_{\mathrm{target}})^2.
\end{array}\]

For small $\delta>0$, we choose 
\[r=\frac{1}{2b}\log\frac2\delta,\ N=\Theta\left(\frac{r^2}{\delta}\right)=\Theta\left(\frac1\delta\left(\log\frac1\delta\right)^2\right).\]
Then, we have
\[D_N(r)\leq\delta\cdot\mathrm{MMD}(q_{\mathrm{source}},p_{\mathrm{target}})^2.\]
Note that 
\[\epsilon = \frac rN = \Theta\left(\frac\delta r\right)=\Theta\left(\frac{\delta}{\log\frac1\delta}\right).\]
Therefore, by \textbf{Lemma} \ref{lemma: lipschitz constant of f}, when $\delta$ is small enough, the Lipschitz constant of each $f_n$ is less than $\frac12$.
\end{proof}

\subsection{Proof of \textbf{Lemma} \ref{lemma: bound Delta 2}}
\begin{proof}
According to \eqref{eq: MMD kernel version} and \eqref{eq: delta},
\[\Delta(q,p;\hat{f}_{\epsilon}) = \mathbb{E}_{z\sim q,x\sim q} (K(z,x)-K(z+\hat{f}_{\epsilon}(z),x+\hat{f}_{\epsilon}(x))) + 2\mathbb{E}_{z\sim q,x\sim p} (K(z+\hat{f}_{\epsilon}(z),x)-K(z,x))\]
There is a closed-form expression for $\Delta_2(q,p;\hat{f}_{\epsilon})$. According to the remainder of multivariate Taylor polynomials, there exist two maps $\xi_1,\xi_2:\mathbb{R}^d\rightarrow(0,1)$ such that
\[\begin{array}{rll}
    \Delta_2(q,p;\hat{f}_{\epsilon}) =
    & \displaystyle ~~ \mathbb{E}_{z\sim q}\mathbb{E}_{x\sim p}\hat{f}_{\epsilon}(z)^{\top}[\nabla_{zz}^2K(z+\xi_1(z)\hat{f}_{\epsilon}(z),x+\xi_1(x)\hat{f}_{\epsilon}(x))]\hat{f}_{\epsilon}(z) & (=:\Delta_2^{(1)}) \\
    & \displaystyle - \mathbb{E}_{z\sim q}\mathbb{E}_{x\sim q}\hat{f}_{\epsilon}(z)^{\top}[\nabla_{zz}^2K(z+\xi_1(z)\hat{f}_{\epsilon}(z),x+\xi_1(x)\hat{f}_{\epsilon}(x))]\hat{f}_{\epsilon}(z) & (=:-\Delta_2^{(2)}) \\
    & \displaystyle - \mathbb{E}_{z\sim q}\mathbb{E}_{x\sim q}\hat{f}_{\epsilon}(z)^{\top}[\nabla_{zx}^2K(z+\xi_2(z)\hat{f}_{\epsilon}(z),x+\xi_2(x)\hat{f}_{\epsilon}(x))]\hat{f}_{\epsilon}(x) & (=:-\Delta_2^{(3)}) \\
    = & \Delta_2^{(1)} - \Delta_2^{(2)} - \Delta_2^{(3)}.
\end{array}\]

First, we bound $|\Delta_2^{(1)} - \Delta_2^{(2)}|$. Define
\[\psi':=(\mathbb{E}_{x\sim p}-\mathbb{E}_{x\sim q})\phi(x+\xi_1(x) \hat{f}_{\epsilon}(x))=\psi(p,q) + \hat{\psi}_{\epsilon}.\]
Since $\phi$ is $L_{\mathrm{feat}}$-Lipschitz and $0<\xi_1(x)<1$, we have
\[\begin{array}{rl}
    \|\hat{\psi}_{\epsilon}\| 
    & \displaystyle \leq \sup_{x\in\mathbb{R}^d} L_{\mathrm{feat}}\xi_1(x)\|\hat{f}_{\epsilon}(x)\| \\
    & \displaystyle \leq \sup_{x\in\mathbb{R}^d} L_{\mathrm{feat}}\epsilon\|J_{\phi}(x)\psi(p,q)\| \\
    & \displaystyle \leq \sup_{x\in\mathbb{R}^d} L_{\mathrm{feat}}\epsilon\sigma_{\max}(J_{\phi}(x))\|\psi(p,q)\| \\
    & \displaystyle \leq \epsilon L_{\mathrm{feat}}\sqrt{B}\|\psi(p,q)\|.
\end{array}\]
Therefore, 
\[\|\psi'\| \leq (1+\epsilon L_{\mathrm{feat}}\sqrt{B})\|\psi(p,q)\|.\]
For any $z',v\in\mathbb{R}^d$,
\[\begin{array}{rl}
    \displaystyle \left| v^{\top}[\nabla_{zz}^2(\phi(z')^{\top}\psi')]v \right| 
    & \displaystyle = \left| \sum_{i=1}^{d_{\phi}}\psi'_i v^{\top}\nabla^2\phi_i(z')v \right| \\
    & \displaystyle \leq \|\psi'\|\|v\|^2\sqrt{\sum_{i=1}^{d_{\phi}} \max\lambda(\nabla^2\phi_i(z'))^2} \\
    & \displaystyle \leq \sqrt{d_{\phi}}C\|\psi'\|\|v\|^2.
\end{array}\]
By letting $z'=z+\xi_1(z)\hat{f}_{\epsilon}(z)$ and $v=\hat{f}_{\epsilon}(z)$, we have
\[\begin{array}{rl}
    |\Delta_2^{(1)}-\Delta_2^{(2)}| 
    & \displaystyle = \left|\mathbb{E}_{z\sim q} v^{\top}[\nabla_{zz}^2(\phi(z')^{\top}\psi')]v\right| \\
    & \displaystyle \leq \epsilon^2\|\psi(p,q)\|^2\sigma^2_{\max}(J_{\phi})\sqrt{d_{\phi}}C\|\psi'\| \\
    & \displaystyle \leq \epsilon^2\|\psi(p,q)\|^3\sqrt{d_{\phi}}BC(1+\epsilon L_{\mathrm{feat}}\sqrt{B}).
\end{array}\]

Next, we bound $\Delta_2^3$. Observe that
\[\Delta_2^{(3)} = - \mathbb{E}_{z\sim q}\mathbb{E}_{x\sim q}\hat{f}_{\epsilon}(z)^{\top}J_{\phi}(z+\xi_2(z)\hat{f}_{\epsilon}(z)) J_{\phi}(x+\xi_2(x)\hat{f}_{\epsilon}(x))^{\top}\hat{f}_{\epsilon}(x).\]
Therefore,
\[\begin{array}{rl}
    |\Delta_2^{(3)}| 
    & \displaystyle \leq \max_{z\in\mathbb{R}^d}\|\hat{f}_{\epsilon}(z)\|^2 \max_{z\in\mathbb{R}^d} \|J_{\phi}(z)\|^2 \\
    & \displaystyle \leq \epsilon^2 \|\psi(p,q)\|^2 \max_{z\in\mathbb{R}^d}\sigma^4_{\max}(J_{\phi}(z)) \\
    & \leq \epsilon^2 \|\psi(p,q)\|^2 B^2.
\end{array}\]

Combining these bounds, we have 
\[|\Delta_2(q,p;\hat{f}_{\epsilon})| \leq \epsilon^2\cdot \mathrm{MMD}(q,p)^2\left(\|\psi(p,q)\|\sqrt{d_{\phi}}BC(1+\epsilon L_{\mathrm{feat}}\sqrt{B})+B^2\right).\]
\end{proof}

\subsection{Proof of \textbf{Theorem} \ref{thm: second order}} \label{appendix: thm 2}
\begin{proof}
Let 
\[q_0=q_{\mathrm{source}}\text{ and }
q_m = (\mathbf{Id}+f_m)\circ\cdots\circ(\mathbf{Id}+f_1)\#q_{\mathrm{source}},\]
where each
\[f_i(z)= \epsilon J_{\phi}(z)\psi(p_{\mathrm{target}},q_{i-1}).\]
Define $\psi_0=\psi(p_{\mathrm{target}},q_0)$ and assume $\|\psi(p_{\mathrm{target}},q_m)\|\leq\|\psi_0\|$ (which we will prove by induction). Note that 
\[\Delta(q_m,p_{\mathrm{target}};f_m)=\mathrm{MMD}(q_m,p_{\mathrm{target}})^2 - \mathrm{MMD}(q_{m+1},p_{\mathrm{target}})^2.\]

According to \textbf{Lemma} \ref{lemma: bound Delta 1} and \textbf{Lemma} \ref{lemma: bound Delta 2}, we have
\[\begin{array}{rl}
    \displaystyle \frac{\Delta(q_m,p_{\mathrm{target}};f_m)}{\mathrm{MMD}(q_m,p_{\mathrm{target}})^2} 
    & \displaystyle \geq 2b\epsilon-\left(\|\psi(p_{\mathrm{target}},q_m)\|\sqrt{d_{\phi}}BC+B^2\right)\epsilon^2-\|\psi(p_{\mathrm{target}},q_m)\|\sqrt{d_{\phi}}B^{\frac32}CL_{\mathrm{feat}}\epsilon^3 \\
    &  \displaystyle \geq 2b\epsilon-\left(\|\psi_0\|\sqrt{d_{\phi}}BC+B^2\right)\epsilon^2-\|\psi_0\|\sqrt{d_{\phi}}B^{\frac32}CL_{\mathrm{feat}}\epsilon^3
\end{array} \]
When 
\[\epsilon\leq\epsilon_{\Delta}=\min\left(\frac{b}{2\left(\|\psi_0\|\sqrt{d_{\phi}}BC+B^2\right)},\sqrt{\frac{b}{2\|\psi_0\|\sqrt{d_{\phi}}B^{\frac32}CL_{\mathrm{feat}}}}\right),\]
we have
\[\frac{\Delta(q_m,p_{\mathrm{target}};f_m)}{\mathrm{MMD}(q_m,p_{\mathrm{target}})^2} \geq b\epsilon.\]

Next, by \textbf{Lemma} \ref{lemma: lipschitz constant of f}, in order to satisfy the Lipschitz condition, we require
\[\epsilon \leq \frac{1}{2\sqrt{d\cdot d_{\phi}}L_{\mathrm{Jac}}\|\psi(p_{\mathrm{target}},q_m)\|}.\]
This is satisfied when we assign
\[\epsilon \leq \epsilon_{\mathrm{Lip}} := \frac{1}{2\sqrt{d\cdot d_{\phi}}L_{\mathrm{Jac}}\|\psi_0\|}.\]

Now, we set 
\[\epsilon = \hat{\epsilon} := \min(\epsilon_{\Delta}, \epsilon_{\mathrm{Lip}}).\]
Then, we have
\[\mathrm{MMD}(q_{m+1},p_{\mathrm{target}})^2 \leq (1-b\hat{\epsilon}) \cdot \mathrm{MMD}(q_m,p_{\mathrm{target}})^2,\]
which also implies $\|\psi(p_{\mathrm{target}},q_{m+1})\|\leq\sqrt{1-b\hat{\epsilon}}\|\psi_0\|\leq\|\psi_0\|$. Finally, in order to satisfy \eqref{eq: goal}, we only need to take the number of residual blocks as 
\[N = \frac{\log\frac1\delta}{\log\frac{1}{1-b\hat{\epsilon}}}.\]

\end{proof}

\end{document}